\documentclass{article}

\PassOptionsToPackage{numbers, compress}{natbib}

\usepackage[final]{nips_2018}

\usepackage[utf8]{inputenc} 
\usepackage[T1]{fontenc}    
\usepackage{hyperref}       
\usepackage{url}            
\usepackage{booktabs}       
\usepackage{amsfonts}       
\usepackage{nicefrac}       
\usepackage{microtype}      
\usepackage{graphicx}
\usepackage{multirow}
\usepackage{boldline}
\usepackage{wrapfig,lipsum,booktabs}

\title{Learning a Multi-Modal Policy via Imitating Demonstrations with Mixed Behaviors}

\author{
  Fang-I Hsiao \\
  National Tsing Hua University\\
  \texttt{fyhsiao25@gapp.nthu.edu.tw} \\
  \And
  Jui-Hsuan Kuo \\
  National Tsing Hua University\\
  \texttt{fionakuo26@gapp.nthu.edu.tw} \\
  \And
  Min Sun \\
  National Tsing Hua University\\
  \texttt{sunmin@ee.nthu.edu.tw} \\
}

\begin{document}

\maketitle

\begin{abstract}
  We propose a novel approach to train a multi-modal policy from mixed demonstrations without their behavior labels.  We develop a method to discover the latent factors of variation in the demonstrations. Specifically, our method is based on the variational autoencoder with a categorical latent variable. The encoder infers discrete latent factors corresponding to different behaviors from demonstrations. The decoder, as a policy, performs the behaviors accordingly. Once learned, the policy is able to reproduce a specific behavior by simply conditioning on a categorical vector. We evaluate our method on three different tasks, including a challenging task with high-dimensional visual inputs. Experimental results show that our approach is better than various baseline methods and competitive with a multi-modal policy trained by ground truth behavior labels.
\end{abstract}

\section{Introduction}
Humans have the ability to perform various behaviors. However, learning an intelligent agent to perform multiple behaviors is still a challenging task. In recent years, reinforcement learning (RL) \cite{sutton1998reinforcement} presents promising results for many applications by optimizing a predefined reward function. As a result, the optimal solution is a single behavior performing the best on this predefine reward function. To extend from one single behavior to various behaviors, we need to explicitly define suitable reward functions corresponding to different behaviors. However, manually defining various reward functions is not intuitive and hence impractical in many environments. 

Imitation learning \cite{argall2009survey,billard2008robot} is an efficient way to learn a policy to perform a task. It alleviates the limitation of defining an appropriate reward function by learning a single behavior directly from expert demonstrations. However, standard approaches for imitation learning are incompetent to learn different behaviors from demonstrations with mixed behaviors. Simply applying imitation learning to such demonstrations will end up learning a policy trying to imitate all behaviors but very likely cannot reproduce any behavior accurately.  A straightforward solution is to add a behavior label to each demonstration \cite{rahmatizadeh2017vision}. However, this needs additional labeling cost and requires the behaviors to be defined in advance. Towards addressing these issues, some previous works propose to automatically recover specific reward functions for different behaviors in the data, which is referred to multi-task inverse reinforcement learning \cite{babes2011apprenticeship,dimitrakakis2011bayesian,hausman2017multi,li2017inferring}. These approaches, however, typically rely on numerous environment interactions, which is not practical for many realistic applications (e.g. robotics).  

Recently, a few approaches \cite{morton2017simultaneous,wang2017robust} apply variational autoencoders (VAEs) \cite{kingma2013auto} to this problem. These methods do not require additional rollouts. VAEs allow the policy to perform different behaviors according to the latent vector representations of demonstrations. These methods, as most VAE-based approaches, employ continuous latent variables with a standard Gaussian distribution as the prior and show some promising results. However, these works need to encode the corresponding demonstration to perform a specific behavior since directly sampling from the Gaussian prior is hard to specify the behavior and may result in sub-optimal performance.


In this work, we propose an approach based on the variational autoencoder with a categorical latent variable that jointly learns an encoder and a decoder. The encoder infers discrete latent factors corresponding to different behaviors from demonstrations. The decoder, as a policy, performs different behaviors according to the latent factors. We propose to use the categorical latent variable to learn the multi-modal policy for two reasons. First, demonstrations with mixed behaviors are inherently discrete in many cases since typically there exist salient differences between behaviors. For example, imagine a robotic arm trying to reach 4 different targets. The demonstrations can be naturally split into 4 categories where each category focuses on one specific target. Thus, using a categorical latent variable is appropriate to represent such behaviors. Second, using the categorical latent variable makes the learned policy more controllable. The categorical latent variable can discover the salient variations in the data and result in simple representations of different behaviors, namely the categories. As a result, each category corresponds to a specific behavior and the learned policy can be controlled to reproduce a behavior by simply conditioning on a categorical vector.

We evaluate our method on three different tasks including a robotic arm trying to reach different targets, a bipedal robot with different moving behaviors, and a challenging car driving task containing multiple driving behaviors with high-dimensional visual inputs. To the best of our knowledge, this is the first VAE-based method for this problem that can scale to tasks with high-dimensional inputs. We also demonstrate that our method can still learn distinct behaviors without the prior knowledge of the number of behaviors in the data.

Our contributions are summarized as the following:
\begin{itemize}
\itemsep=0pt
\item We propose an approach based on the variational autoencoder (VAE) with a categorical latent variable to imitate multiple behaviors from mixed demonstrations without behavior labels.

\item We demonstrate our method is applicable to several tasks, including a challenging task with high-dimensional visual inputs.

\item We show the categorical latent variable can discover distinct behaviors without the prior knowledge of the number of behaviors in the data.

\end{itemize}

\section{Related work}
Imitation learning considers the problem of learning skills from demonstrations. Two main approaches of imitation learning are behavior cloning \cite{pomerleau1991efficient} and inverse reinforcement learning \cite{abbeel2004apprenticeship,ho2016generative,levine2011nonlinear,Ng:2000:AIR:645529.657801,ziebart2008maximum}. Behavior cloning directly mimics expert demonstrations by supervised learning from states to actions without interacting with the environment, providing an efficient way to learn how to perform a task. In contrast, inverse reinforcement learning is aimed to seek a reward function that can explain the behavior shown in demonstrations. However, these methods typically assume the demonstrations consist of a single behavior.

Multi-task inverse reinforcement learning \cite{dimitrakakis2011bayesian,babes2011apprenticeship} aims to learn from demonstrations with multiple behaviors by inverse reinforcement learning. In \cite{dimitrakakis2011bayesian}, the authors propose a Bayesian approach for inferring the intention of an agent. In \cite{babes2011apprenticeship}, the authors propose an approach based on the EM algorithm that clusters observed trajectories by using inverse reinforcement learning methods to infer the intention for each cluster. In contrast to previous methods, recent work on multi-task inverse reinforcement learning has adopted the generative adversarial imitation learning (GAIL) algorithm \cite{li2017inferring,hausman2017multi}. The authors propose a similar framework that extends GAIL to incorporate a component that maximizes mutual information between latent variables and state-action pairs. 

Multi-task inverse reinforcement learning typically needs an inefficient procedure of taking additional rollouts, which is hard to be applied to many realistic applications. Recently, a stochastic neural network framework \cite{tamar2018imitation} is proposed to learn a multi-modal policy without additional rollouts by selecting the sampled intention with the lowest error for updating network parameters. Similarly, our approach does not rely on any additional rollouts and directly learn multiple behaviors from demonstrations.

While these methods achieve some promising results, some works explore another approach which is often employed to learn a latent variable generative model called variational auto-encoders (VAEs) \cite{morton2017simultaneous,wang2017robust}. To perform a specific behavior, these methods need to condition on its corresponding demonstration. Our approach, in comparison, can learn categorical representations for distinct behaviors, resulting in a policy that can perform a specific behavior by conditioning on a categorical vector.

\section{Preliminaries}
\label{pre}
In this section, we formally define the notations as well as describe some background concepts about imitation learning and variational autoencoders.
\subsection{Imitation learning} 
Let $S$ represent the state space, $A$ represent the action space. In imitation learning setting, we assume that we have been provided a set of demonstrations which consists of $N$ trajectories $\{\zeta_i\}_{i=1}^N$. Each trajectory $\zeta_i$ is composed of a sequence of state-action pairs $\zeta_i = \{s_1^i, a_1^i, ..., s_T^i, a_T^i\}$ where $s_t^i \in S, a_t^i \in A$ denote the state and action respectively at time $t$. For the remainder of this paper, we drop the notation $i$ and abbreviate the state sequence $s_1, ..., s_t$ as $s_{1:T}$ and the action sequence $a_1, ..., a_t$ as $a_{1:T}$ for simplicity. Let $\pi_\theta$ denote a policy that defines the distribution over actions given states. The goal of imitation learning is to learn a policy from demonstrations such that it can reliably perform a task. To learn the policy, we can maximize the log likelihood: 
\begin{equation}
\mathcal{L}(\theta;\zeta) = \;\sum_{i=1}^{N}\log\pi_\theta(a_{1:T}|s_{1:T}) = \sum_{i=1}^{N}\sum_{t=1}^{T}\log\pi_\theta(a_t|s_t).
\end{equation}
The second equality is a consequence of a standard assumption that the policy is memory-less, which implies $\pi_\theta(a_{1:T}|s_{1:T}) = \prod_{t=1}^{T}\pi_\theta(a_t|s_t)$.

\subsection{Variational autoencoder}
The variational autoencoders (VAEs) \cite{kingma2013auto} are a kind of generative model with certain types of latent variables. The generative process is composed of two steps, where first a latent variable $z$ is sampled from some prior distribution $p(z)$ and then the data $x$ is generated from some conditional distribution $p_\theta(x|z)$. Often the likelihood of the data is intractable. Therefore, VAEs introduce a encoder $q_\phi(z|x)$ to approximate the true posterior $p(z|x)$ and optimize the variational lower bound of the log-likelihood:

\begin{equation}
\log p(x) \geq \mathcal{L}(\theta,\phi;x,z) = \mathrm{E}_{z \sim q_\phi(z|x)} \left[\log p_\theta(x|z)\right] -D_{KL}(q_\phi(z|x) || p(z)).
\end{equation}

We can combine additional information $y$ to generate the data $x$, and optimize the variational lower bound of the conditional log-likelihood:
\begin{equation}
\log p(x|y) \geq \mathcal{L}(\theta,\phi;x,y,z) = \;\mathrm{E}_{z \sim q_\phi(z|x,y)} \left[\log p_\theta(x|y,z)\right] -D_{KL}(q_\phi(z|x,y) || p(z|x)).
\end{equation}
This class of models is referred to as conditional variational autoencoders (CVAEs) \cite{NIPS2015_5775}.

\begin{figure*}
  \centering
  \includegraphics[width=\textwidth]{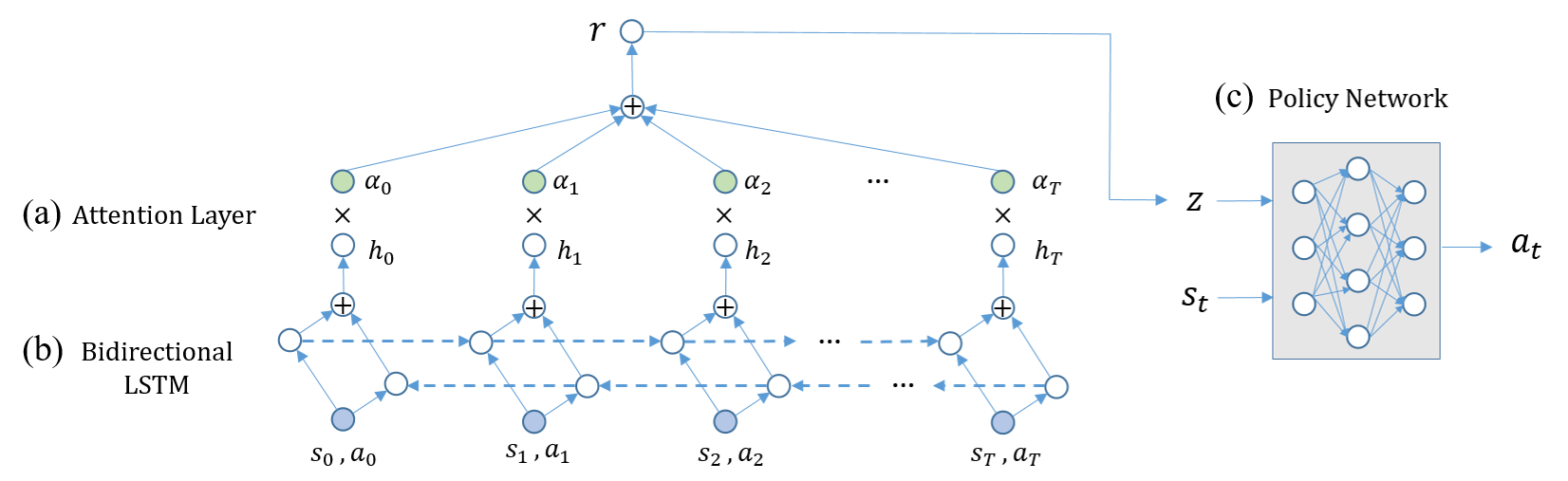}
  \caption{Illustration of our method. During training, each trajectory is encoded by a bidirectional LSTM with an attention mechanism. The approximated categorical posterior $q(z|s_{1:T}, a_{1:T})$ is parameterized by feeding the representation of the whole trajectory through a fully connected layer with a softmax function. The policy takes a latent variable $z$ which is sampled from the posterior and a state to generate an action at each time step. At test time, the policy can perform different behaviors by directly sampling $z$ from a categorical prior distribution where each category corresponds to a specific behavior.}
  \centering
  \label{fig:schematic}
\end{figure*}

\section{Our method}
In this section, we first construct our method based on conditional variational autoencoders (CVAEs) for imitation learning. Next, we describe our architecture with an attention mechanism. Finally, we present a categorical reparameterization trick which allows us to train the variational autoencoder with categorical latent variables.  The overall architecture of our method is shown in Figure \ref{fig:schematic}.

\subsection{Variational autoencoder for behavior cloning}
In order to learn a multi-modal policy from demonstrations with mixed behaviors, we formulate a probabilistic graphical model using CVAEs. Given a sequence of $N$ trajectories, we assume each trajectory $\zeta$ is generated from an unobserved latent variable $z$ and $z$ is the same throughout the whole trajectory. Hence, the conditional generative process is as follows: given states $s_{1:T}$, $z$ is sampled from the prior distribution $p(z|s_{1:T})$, and actions $a_{1:T}$ are generated from the distribution $\pi_\theta(a_{1:T}|s_{1:T}, z)$. We can factorize it to $\prod_{t=1}^{T}\pi_\theta(a_t|s_t, z)$ with a memory-less policy, which means at each time $t \in \{1, ..., T\}$, the action $a_t$ is generated from only the state $s_t$ and $z$. Note that the prior distribution is modulated by $s_{1:T}$, which is infeasible at test time since we are unable to infer $z$ using future states $s_{t+1:T}$ at time $t$. This constraint, however, can be relaxed by making an assumption that the latent variables are statistically independent of the states \cite{doersch2016tutorial,walker2016uncertain}. That is, we assume $p(z|s_{1:T}) = p(z)$ where $p(z)$ is some assumed prior distribution, and thus we can directly sample $z \sim p(z)$ to perform a task instead of sampling from $p(z|s_{1:T})$ at test time.

We train our CVAE by maximizing the conditional log-likelihood $\pi_\theta(a_{1:T}|s_{1:T})$. We optimize its variational lower bound by introducing an encoder $q_\theta(z|s_{1:T}, a_{1:T})$ that can approximate the true posterior distribution $p(z|s_{1:T}, a_{1:T})$. To be specific, we maximize the following objective function:
\begin{equation}
\mathcal{L}(\theta,\phi;\zeta,z) = \mathrm{E}_{z \sim q_\phi(z|s_{1:T}, a_{1:T})} \left[\sum_{t=1}^{T}\log\pi_\theta(a_t|s_t,z)\right] -D_{KL}(q_\phi(z|s_{1:T}, a_{1:T}) || p(z)).
\end{equation}
Typically $p(z)$ is assumed to be the standard Gaussian distribution $\mathcal{N}(\mathbf{0}, \mathbf{I})$. We propose, however, to employ a categorical distribution to learn behavior-level representations. 

Intuitively, since the different behaviors typically have salient differences in terms of trajectories, we can learn categorical representations that correspond to distinct classes of behaviors from their trajectories. In our formulation, the latent variable $z$ contains high-level information of the whole trajectory. Therefore, employing the categorical prior distribution enables our model to discover the variation between trajectories and learn categorical representations for different behaviors. Consequently, the policy trained by the categorical latent variable can perform different behaviors by simply changing the categories, where each category corresponds to a specific behavior.

\subsection{Model architecture}
Our recognition model $q_\phi(z|s_{1:T}, a_{1:T})$ uses a bi-directional LSTM \cite{schuster1997bidirectional}, which encodes the whole trajectory to obtain its context representations. The bidirectional LSTM (Figure \ref{fig:schematic}(b)) combines a forward $\overrightarrow{LSTM}$ which processes the trajectory from $t = 1$ to $T$ and a backward $\overleftarrow{LSTM}$ which processes the trajectory from $t = T$ to $1$:

\[\overrightarrow{h_t} = \overrightarrow{LSTM}(s_t, a_t, \overrightarrow{h_{t-1}}), \qquad
\overleftarrow{h_t} = \overleftarrow{LSTM}(s_t, a_t, \overleftarrow{h_{t+1}}),\]
and we obtain a hidden state $h_t$ by using element-wise sum to combine the forward state $\overrightarrow{h_t}$ and the backward state $\overleftarrow{h_t}$. 

To produce the final representations of the whole trajectory, we propose the attention mechanism \cite{bahdanau2014neural,yang2016hierarchical,duan2017one} to extract the pairs that are important to the meaning of the trajectory. Specifically, the attention mechanism calculates a vector of weights used for a linear combination of all hidden states:
\begin{equation}
u_t = \tanh(W_wh_t + b_w),
\end{equation}
\begin{equation}
\alpha_t = \frac{\exp(u_t^Tu_w)}{\sum_{t'=1}^{T}\exp(u_{t'}^Tu_w)},
\end{equation}
A simple schematic of attention mechanism is shown in (Figure \ref{fig:schematic}(a)).
We first feed $h_t$ through a one-layer MLP, then a high-level representation of a
query $u_w$ is introduced to measure the importance of
each pair and calculate the importance weights $\alpha_t$ by a softmax function. The query vector $u_w$ is randomly initialized and learned jointly. The representations of the trajectory are then calculated by a simple linear combination:
\begin{equation}
r = \sum_{t=1}^{T} \alpha_t h_t.
\end{equation}
This attention mechanism aims to emphasize the informative pairs that can better represent the trajectory, which is similar to that used in document classification \cite{yang2016hierarchical}. The approximated posterior distribution is obtained by feeding $r$ through a one-layer MLP with a softmax function. Finally, we concatenate $z$, which is sampled using a reparameterization trick described in the following section, and a state as the input of a standard MLP policy (Figure \ref{fig:schematic}(c)) to generate an action.

\subsection{Reparametrization of discrete latent variables}
To train our conditional variational autoencoder with categorical latent variables, we employ a categorical reparameterization trick with Gumbel-Softmax distribution \cite{jang2016categorical,maddison2016concrete}. Gumbel-Softmax distribution is a continuous distribution that can approximate sampling from a categorical distribution. Let the approximated posterior $q_\phi(z|s_{1:T}, a_{1:T})$ represent a categorical distribution over $k$ classes with class probability $\lambda_1, ..., \lambda_k$ and $z \sim q_\phi(z|s_{1:T}, a_{1:T})$ is represented by a $k$-dimensional one-hot vector. Sampling $z$ according to this distribution can be replaced with the Gumbel-Max trick which draws samples $z$ according to:
\begin{equation}
z = \mathrm{one\_hot} \left(\mathop{\mathrm{argmax}}\limits_{i} g_i + \log\lambda_i\right),
\end{equation}
where $g_i$ are i.i.d samples drawn from a standard Gumbel distribution.

Because argmax is a non-differentiable operation, we use the softmax function to approximate it and relax $z$ to a continuous random variable $z'$ which can be expressed as:
\begin{equation}
z'_i = \frac{\exp{((g_i + \log\lambda_i)/\tau)}}{\sum_{j=1}^{k}\exp((g_j + \log\lambda_j)/\tau)} \quad \mathrm{for} \;i = 1, ...,\,k,
\end{equation}
where $\tau$ is the temperature parameter. As $\tau \rightarrow 0$, Gumbel-Softmax distribution approaches a categorical distribution while as $\tau \rightarrow \infty$, it becomes a uniform distribution.

We employ the straight-through variation of Gumbel-Softmax distribution \cite{jang2016categorical} to sample the output of the approximated posterior distribution. That is, we use $z$ as the input of our policy but back-propagate gradients according to its continuous relaxation $z'.$ Using straight-through variation during training is more consistent with testing since at test time, we simply input different one-hot vectors to our policy to perform different behaviors.

\section{Experiments}
Our primary goal of experiments is to justify whether our method can automatically discover discrete latent factors of variation in demonstrations with multiple behaviors, and learn a multi-modal policy that can perform different behaviors by conditioning on a categorical vector. 
We evaluate our method on three different tasks (Figure \ref{fig:envs}), which includes simulated robotic tasks, and a challenging car driving task with visual inputs:

\begin{itemize}
	\item \textbf{FetchReach}: The Fetch environments from OpenAI gym \cite{1606.01540} are based on the 7-DoF robotic arm.   The goal of this task is to reach a target position.  We place 4 targets randomly in 4 quadrants of the table.  The demonstrations include 4 kinds of behaviors, where each behavior reaches the target in its corresponding quadrant. 
	\item \textbf{Walker-2D}: The Walker-2D environment from Deepmind Control Suite \cite{tassa2018deepmind} is based on a 6-DoF bipedal robot.  We use a set of demonstrations which consist of 8 different moving behaviors separated by speeds.
	\item \textbf{TORCS}: TORCS is a car racing simulator \cite{TORCS} which provides high-dimensional visual inputs. In this environment, our policy receives only raw visual inputs as the states and generates two continuous actions made up of \textit{steering} and \textit{acceleration}. We consider two experimental setups. One is keeping the car to the left or right, which we refer to \textbf{\textit{direction}}, and the other is to drive the car at certain speeds, which we refer to \textbf{\textit{speed}}.

\begin{figure}[t]
\includegraphics[clip=True, width=\linewidth]{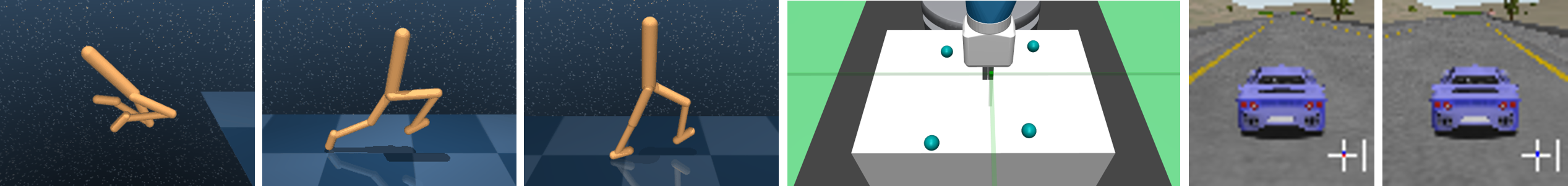}
\caption{Our experiment environments. \textbf{Left}: Examples of different behaviors in Walker-2D. \textbf{Middle}: FetchReach with 4 random targets. \textbf{Right}: Different driving behaviors in TORCS \textbf{\textit{direction}}.}
\label{fig:envs}
\end{figure}

\end{itemize}

\subsection{Reaching multiple targets}
In the FetchReach experiment, we place the targets randomly at 4 quadrants of the table.  The goal of each policy is to reach the target at its corresponding quadrant. This experimental setting provides a concrete example of demonstrations that are inherently discrete with multiple distinct behaviors.  This experiment aims to justify whether or not our method can automatically learn multiple behaviors from mixed demonstrations. We compare our approach to the following methods: 
\begin{itemize}
	\item \textbf{BC without labels}: We train a behavior cloning policy with all demonstrations as a baseline.
	\item \textbf{BC with labels}: The architecture is as same as the decoder of our model, while the input is a state concatenated with a one-hot vector of a given behavior label.  This model is designed to provide an upper bound on performance since it has knowledge of behavior labels.
	\item \textbf{Gaussian VAE}:  We use the same architecture as our method but change the prior distribution to a unit Gaussian distribution. We set the dimension of the latent variable z to 4, which is as same as our method.  Since the latent vectors of Gaussian cannot be directly designated as our method, we evaluate the performance of Gaussian VAE by 2 kinds of latent vectors. One is randomly sampled from a prior distribution, and the other is encoded from trajectories that successfully reach different targets.
\end{itemize}

\begin{table*}
	\newcolumntype{P}[1]{>{\centering\arraybackslash}p{#1}}
	\centering
	\caption{FetchReach experiment results for our approach and baselines.}
    \begin{tabular}{l c c c c c c}
    \\
    \hlineB{2}
    Approach & Success Rate    & z & target 0  & target 1 & target 2 & target 3  \\ \hline

  BC w/o labels  & 33.25\%  & N/A & 98 & 18 & 6 & 11 \\ \hline
  
   \multirow{6}{*}{Gaussian VAE}  &  \multirow{2}{*}{49.5\%} & \multirow{2}{2cm}{Sampled from the prior} & \multirow{2}{*}{54} & \multirow{2}{*}{58} & \multirow{2}{*}{42} & \multirow{2}{*}{44} \\  &&&&&& \\ \cline{2-7}
   & \multirow{4}{*}{77.25\%}  & \multirow{4}{2cm}{Encoded from the trajectory} & 87 & 0 & 0 & 0  \\ 
      &&& 0 & 84 & 0 & 0 \\ 
      &&& 0 & 0 & 79 & 0 \\ 
      &&& 0 & 0 & 0 & 59 \\ \hline
      
   \multirow{4}{*}{Ours}  & \multirow{4}{*}{ \textbf{98\%}}  & [1,0,0,0] & 0 & 0 & 96 & 0 \\ 
      && [0,1,0,0] & 0 & 100 & 0 & 0 \\ 
      && [0,0,1,0] & 0 & 0 & 0 & 98 \\ 
      && [0,0,0,1] & 98 & 0 & 0 & 0 \\ \hline
      
   \multirow{4}{*}{BC w/ labels}  & \multirow{4}{*}{100\%} & [1,0,0,0] & 100 & 0 & 0 & 0 \\ 
      && [0,1,0,0] & 0 & 100 & 0 & 0 \\ 
      && [0,0,1,0] & 0 & 0 & 100 & 0 \\ 
      && [0,0,0,1] & 0 & 0 & 0 & 100 \\ \hlineB{2}
      
    \end{tabular}\hspace{10pt}
    
	\label{table:fetchreach_result}
\end{table*}

For training data, we train the expert policy by HER \cite{andrychowicz2017hindsight} reaching to different targets and collect 600 trajectories for each target.

For evaluation, we randomly choose 100 configurations for each behavior, total 400 configurations.  We evaluate the overall success rate from the total targets reached by the gripper and the count of success reach of different latent vectors.  The result is shown in Table.\ref{table:fetchreach_result}.  Not surprisingly, BC without labels is unable to perform four behaviors. It nearly collapses to a single mode and results in a low success rate.  We can observe that Gaussian VAE with latent vectors sampled from prior has better performance than BC w/o labels.  However, the success rate is still not satisfactory.  We consider that it is because the random sampled latent vector may lead the gripper to one position on the table, yet it is not suitable for the target position in the configuration.  For Gaussian VAE with latent vectors encoded from trajectories, something noteworthy is that the success rate is lower than expected since we provide the successful trajectories for the model.  We surmise that the reason for the poor performance is that the model has never seen the test configurations and the trajectories before. Therefore, it may not map the given trajectories to appropriate latent vectors. Finally, we can see that our method outperforms Gaussian VAE, and the performance of our method is competent to BC with labels. It can not only distinguish between different behaviors but also reach targets precisely by only conditioning on a one-hot vector that corresponds to a fixed behavior.

\begin{figure}[t]
\includegraphics[clip=True, width=\textwidth]{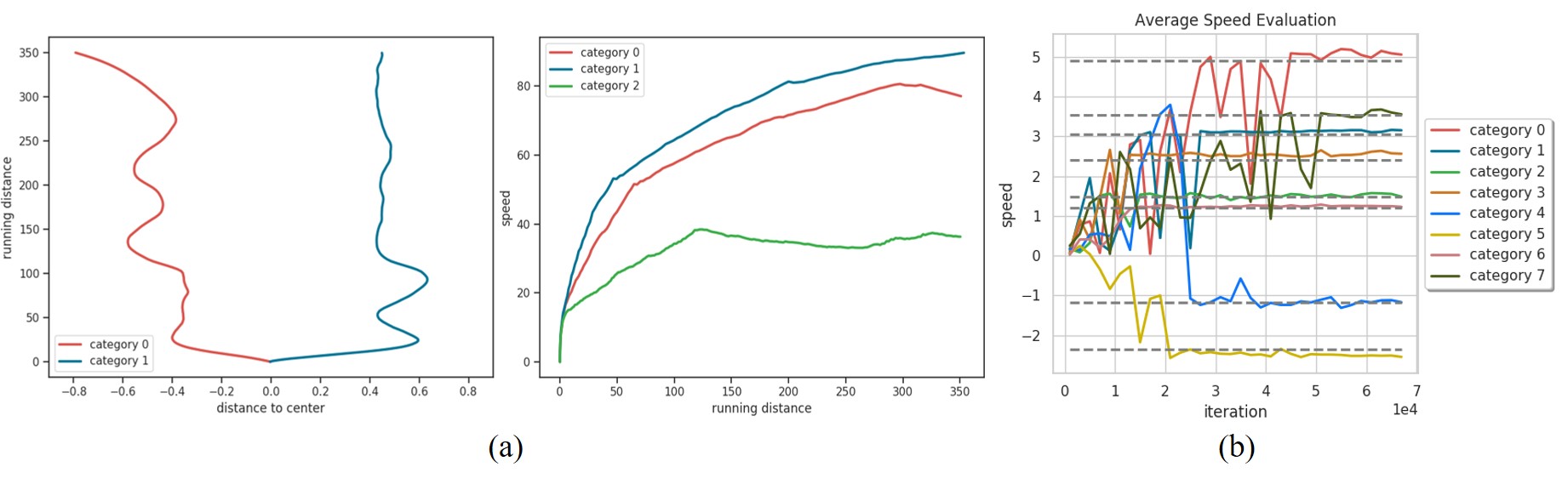}
\caption{Results of TORCS and Walker-2D. (a) For the TORCS experiment, we plot the trajectories generated from our policy. \textbf{Left}: Results of \textbf{\textit{direction}}. The x-axis represents the distance between the car and the track axis, and the y-axis represents the distance from the start. \textbf{Right}: Results of \textbf{\textit{speed}}. the x-axis represents the distance from the start, and the y-axis represents the speed. (b) For Walker-2D, we evaluate our policy conditioning on different latent vectors during training. The dashed lines represent the average speeds of demonstrations of different expert policies.}
\label{fig:walker_reward}
\end{figure}

\subsection{Moving behaviors}
In the Walker-2D experiment, the goal is to justify whether our approach can learn to separate and imitate different moving behaviors. We use PPO \cite{schulman2017proximal} to train expert policies reaching 8 different target speeds and collect 100 trajectories from each policy. Since the policies are trained by reaching different speeds without any constraint on the robot's movement, different behaviors may consist of similar state-action pairs. However, the behaviors are still distinct in terms of trajectories and we expect our approach can work well. We consider this experiment more challenging than FetchReach.  

Since the behaviors are separated by speeds, we present speeds over training iterations to demonstrate the development of different latent vectors. The result is shown in Figure \ref{fig:walker_reward} (b). Our model can gradually separate 8 different behaviors from mixed data and successfully imitate each behavior. The policy can reach the same speeds as the expert policies by simply conditioning on different one-hot vectors. This result indicates that our method can still perform well even learning from complex demonstrations.

\subsection{Driving behaviors}
In the TORCS experiment, our goal is to evaluate if our approach can generalize to tasks with high-dimensional inputs. We design a heuristic agent to collect demonstrations. For \textbf{\textit{direction}}, there are two behaviors: keeping to the left or right. For \textbf{\textit{speed}}, there are three behaviors: driving at speed [40, 60, 80]. We collect 30 trajectories for each behavior. 

Directly learning from raw visual inputs is challenging as the model is required to extract meaningful visual features and identify the meaning of different trajectories simultaneously. Besides, since we extract the information from a whole trajectory during training, it is limited by memory sizes and also brings computational inefficiency especially when dealing with long trajectories. Hence, we adopt a transfer learning approach.
In particular, We extract image features from the average pooling layer before the final classifier of GoogLeNet \cite{szegedy2015going} with the dimension of $1 \times 1024$ as the states of the trajectory. We do not feed any additional information such as current speeds into our model, forcing it to learn to distinguish different behaviors from only image features.

We visualize our results in Figure \ref{fig:walker_reward} (a). We can observe that our method can clearly separate different behaviors for both \textbf{\textit{direction}} and \textbf{\textit{speed}} even using high-dimensional visual inputs, demonstrating the ability to generalize to tasks with only visual inputs. 

\begin{figure*}[t]
  \centering
  \includegraphics[width=\textwidth]{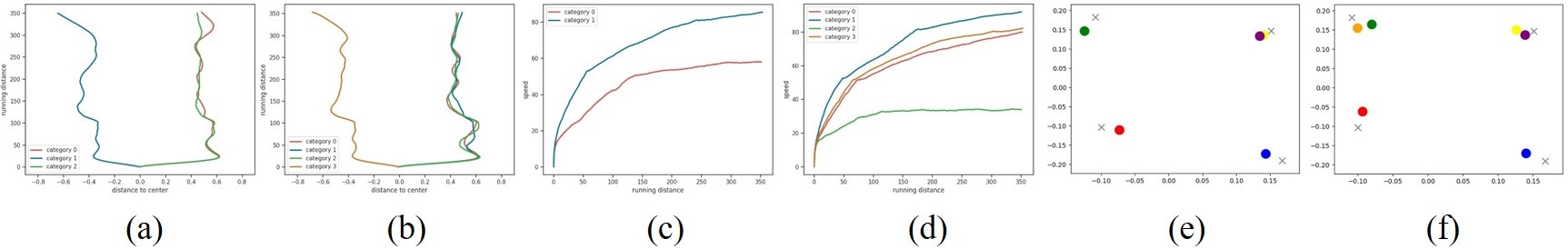}

  \caption{(a)(b) Results of \textbf{\textit{direction}} with the number of categories greater than the number of behaviors. (c)(d) Results of \textbf{\textit{speed}} with the number of categories less (c) and greater (d) than the number of behaviors. (e)(f) Results of FetchReach experiment with the number of categories greater than the number of behaviors. Circles of different colors are the final positions of the gripper with 5 categories (e) and 6 categories (f).  Crosses are the positions of targets.}
  \centering
  \label{fig:without_prior}
\end{figure*}


\subsection{Learning without prior knowledge}
We demonstrate that our model can successfully learn multiple behaviors without the prior knowledge of the number of behaviors in the data. Learning without the prior knowledge is important for many practical applications since in many cases we are unable to know variations in the data in advance. We conduct this experiment in the TORCS and FetchReach environments. We make the number of categories different from the number of behaviors in the demonstration, and test if our model can still separate different behaviors. We show the results in Figure \ref{fig:without_prior}. (a)(b) are the results of \textbf{\textit{direction}} experiment in TORCS with 3 and 4 categories. (c)(d) are the results of \textbf{\textit{speed}} experiment in TORCS with 2 and 4 categories.  (e)(f) are the results of the FetchReach experiment with 5 and 6 categories.  As expected, we can see that our model groups similar behaviors into the same category if we employ fewer categories (Figure \ref{fig:without_prior} (c)). However, we can observe that if the number of categories is greater than the number of behaviors in the data (Figure \ref{fig:without_prior} (a)(b)(d)(e)(f)), our model can still find all behaviors in the demonstrations and imitate each behavior. Our model will first discover the inter-class variation and then try to seek the intra-class variation since we find it separates the same behavior into different groups. These results suggest that our method can be used to learn a multi-modal policy from demonstrations that contain distinct unknown behaviors.


\section{Conclusion}
We present a method to learn a multi-modal policy from demonstrations with mixed behaviors. The presented approach is based on the variational autoencoder with a categorical latent variable which learns the representations corresponding to different behaviors in demonstrations. Our experimental results show our method can work on a variety of tasks, including a challenging task with high-dimensional visual inputs. We also show that using the categorical variable can automatically discover all distinct behaviors without prior knowledge of the number of behaviors in demonstrations.  In the future, we plan to scale up our method to more realistic demonstrations which contain a larger number of complex behaviors.

\bibliographystyle{abbrv}
\bibliography{ref.bib}
\end{document}